\documentclass[10pt,twocolumn,letterpaper]{article}

\usepackage{iccv}
\usepackage{times}
\usepackage{epsfig}
\usepackage{graphicx}
\usepackage{amsmath}
\usepackage{amssymb}
\usepackage{booktabs}
\usepackage{multirow}
\usepackage{multicol}
\usepackage{color, colortbl}
\usepackage{caption}
\usepackage{subcaption}
\usepackage{bm}
\usepackage{pifont}
\newcommand{\xmark}{\ding{55}}%
\definecolor{Gray}{gray}{0.9}
\usepackage[normalem]{ulem}

\newcommand{\sprod}[2]{\left\langle #1,#2 \right\rangle}

\renewcommand{\vec}[1]{\bm{#1}}

\newcommand{\ours}{Soft Neighborhood Density\xspace}
\newcommand{\oursv}{SND\xspace}

\usepackage[pagebackref=true,breaklinks=true,letterpaper=true,colorlinks,bookmarks=false]{hyperref}

\iccvfinalcopy 

\def\iccvPaperID{6193} 

\ificcvfinal\fi

\begin{document}

\title{Tune it the Right Way: Unsupervised Validation of Domain Adaptation  \\
via \ours}

\author{%
  Kuniaki Saito$^{1}$, Donghyun Kim$^{1}$, Piotr Teterwak$^{1}$, Stan Sclaroff$^{1}$, Trevor Darrell$^{2}$, and Kate Saenko$^{1,3}$\\
  $^{1}$Boston University, $^{2}$University of California, Berkeley, $^{3}$MIT-IBM Watson AI Lab\vspace{.7em}\\
\texttt{[keisaito,donhk,piotrt,sclaroff,saenko]@bu.edu, trevor@eecs.berkeley.edu}}

\maketitle
\ificcvfinal\thispagestyle{empty}\fi


\begin{abstract}
Unsupervised domain adaptation (UDA) methods can dramatically improve generalization on unlabeled target domains. However, optimal hyper-parameter selection is critical to achieving high accuracy and avoiding negative transfer. Supervised hyper-parameter validation is not possible without labeled target data, which raises the question: How can we validate unsupervised adaptation techniques in a realistic way? We first empirically analyze existing criteria and demonstrate that they are not very effective for tuning hyper-parameters. Intuitively, a well-trained source classifier should embed target samples of the same class nearby, forming dense neighborhoods in feature space. Based on this assumption, we propose a novel unsupervised validation criterion that measures the density of soft neighborhoods by computing the entropy of the similarity distribution between points. Our criterion is simpler than competing validation methods, yet more effective; it can tune hyper-parameters and the number of training iterations in both image classification and semantic segmentation models. The code used for the paper will be available at \url{https://github.com/VisionLearningGroup/SND}.
\end{abstract}

 \vspace{-5mm}
\section{Introduction}
\vspace{-2mm}
Deep neural networks can learn highly discriminative representations for visual recognition tasks~\cite{imagenet, simonyan2014very, krizhevsky2012imagenet, faster, maskrcnn},
but do not generalize well to out-of-domain data~\cite{tzeng2014deep}. To improve performance on a new target domain, Unsupervised Domain Adaptation (UDA) aims to transfer representations from a label-rich source domain without additional supervision. 
Recent UDA methods primarily achieve this through unsupervised learning on the target domain, \eg, by minimizing the feature distribution shift between source and target domains~\cite{ganin2014unsupervised, long2015learning, sun2015return}, classifier confusion~\cite{jin2020minimum}, clustering~\cite{saito2020universal}, and pseudo-label based methods~\cite{zou2018unsupervised}.
\begin{figure}
    \centering
    \includegraphics[width=\linewidth]{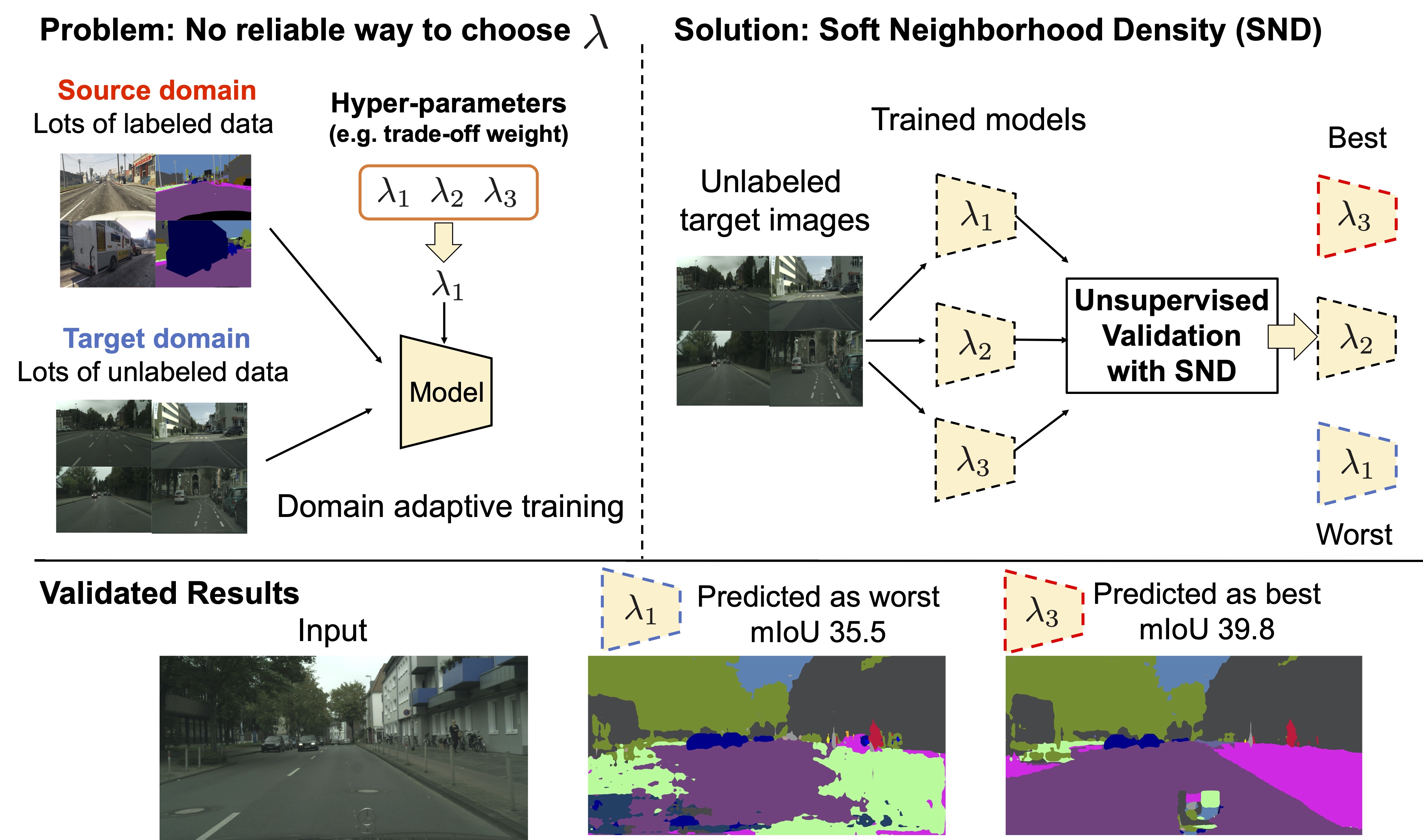}
    \vspace{-6mm}
    \caption{{\small 
    In unsupervised domain adaptation, validation is a significant and unsolved issue. Performance can be sensitive to hyper-parameters, yet no reliable validation criteria have been presented. 
    In this work, we provide a new criterion, \textit{\oursv}, to select proper hyper-parameters for model validation in UDA. The example shows validation of the ADVENT~\cite{vu2018advent} segmentation model.}}
    \label{fig:fig1}
    \vspace{-15pt}
\end{figure}
Promising adaptation results have been demonstrated on  image classification~\cite{long2017conditional,zou2019confidence, chen2019transferability,xu2019larger,tang2020unsupervised},
 semantic segmentation~\cite{hoffman2016fcns} and object detection~\cite{dafaster} tasks.

However, adaptation methods can be highly sensitive to hyper-parameters and the number of training iterations. For example, the adversarial alignment approach is popular in semantic segmentation~\cite{hoffman2016fcns,tsai2018learning,vu2018advent}, but can fail badly without careful tuning of the loss trade-off weight as shown in Fig.~\ref{fig:fig1}. 
In addition, many methods have other kinds of hyper-parameters, \eg, defining the concentration of clusters~\cite{saito2020universal}, the confidence threshold on target samples~\cite{zou2018unsupervised}, etc.
Therefore validation of hyper-parameters is an important problem in UDA, yet it has been largely overlooked. In UDA, we assume not to have access to labeled target data, thus,  hyper-parameter optimization (HPO) should be done without using target labels. 
In this paper, we would like to pose a question crucial to making domain adaptation more practical: \textit{How can we reliably validate adaptation methods in an unsupervised way?}
 
Unsupervised validation is very challenging in practice, thus many methods do HPO in an ineffective, or even unfair, way. 
Evaluating accuracy (risk) in the source domain is popular~\cite{ganin2014unsupervised,sun2015return,cao2018partial, sugiyama2007covariate,long2017conditional,cao2018partial,jin2020minimum,wang2019transferable,you2019towards}, but this will not necessarily reflect success in the target domain. Using the risk of the target domain ~\cite{shrivastava2017learning,bousmalis2017unsupervised,bousmalis2016domain,ghifary2016deep,saito2017asymmetric,shu2018dirt,russo2018source} contradicts the assumption of UDA. Many works~\cite{tzeng2015simultaneous,sun2016deep,tsai2018learning,murez2018image,zou2018unsupervised,zhang2018collaborative,volpi2018adversarial,vu2018advent} do not clearly describe their HPO method. 

 To the best of our knowledge, no comprehensive study has compared validation methods across tasks and adaptation approaches in a realistic evaluation protocol. Our first contribution is to empirically analyze these existing criteria and demonstrate that they are not very effective for HPO. This exposes a major barrier to the practical application of state-of-the-art unsupervised domain adaptation methods.

\begin{figure}
    \centering
    \includegraphics[width=\linewidth]{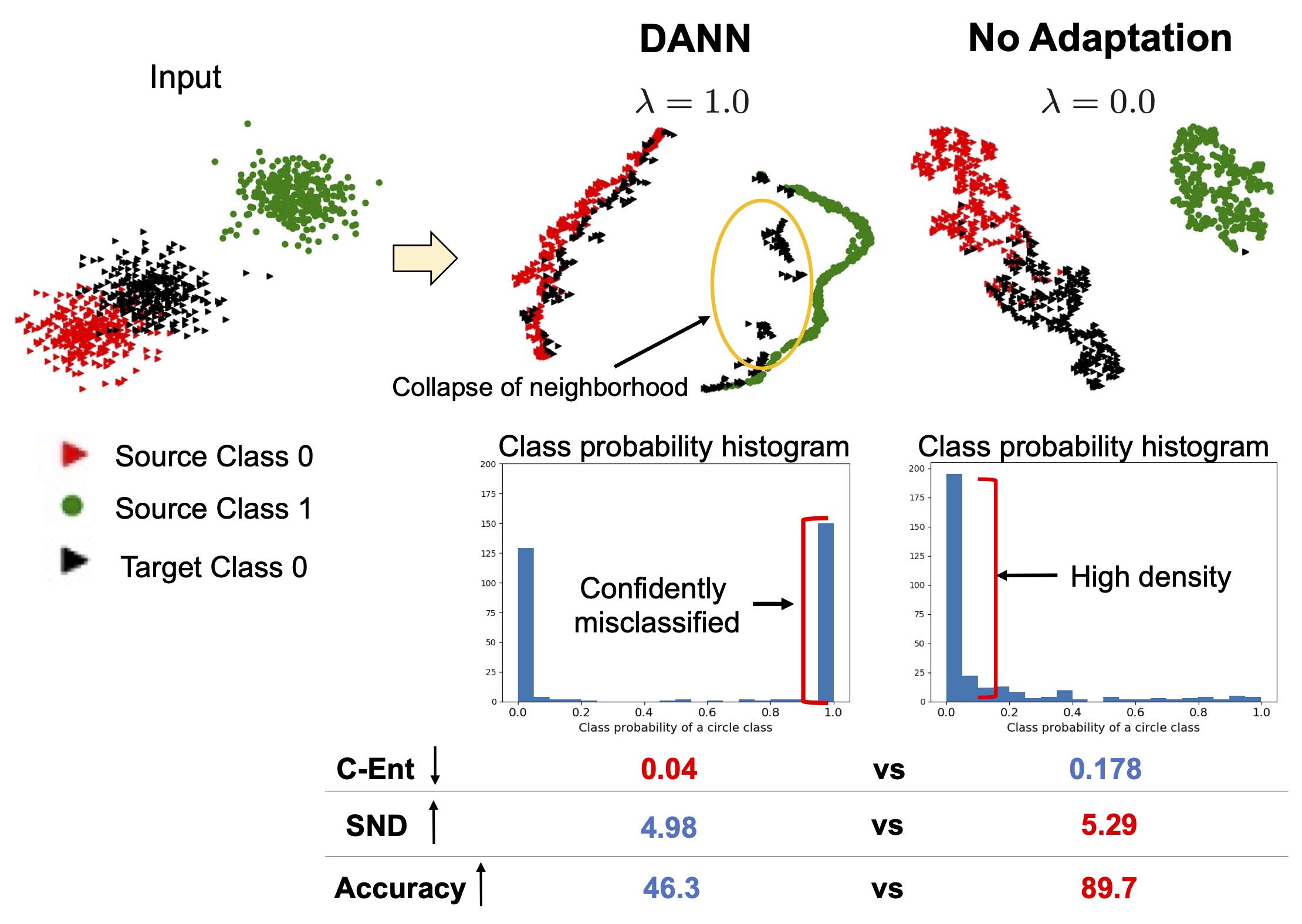}
    \vspace{-7mm}
    \caption{
    {\small{{\bf An illustration of C-Ent failing to detect the collapse of target neighborhood structure.} This is the case where an adapted model ($\lambda=1.0$) confidently misclassifies target samples and low entropy (C-Ent) cannot select a good model. The model incorrectly changes the relative distances between target samples. \oursv can select a better model since it can consider how well the neighborhood structure is preserved.}}}
    \label{fig:idea}
    \vspace{-10pt}
\end{figure}

To tackle this problem, we first revisit an unsupervised validation criterion based on classifier entropy, C-Ent. If the classification model produces confident and low-entropy outputs on target samples, then the target features are discriminative and the prediction is likely reliable. 
Morerio \etal~\cite{morerio2017minimal} propose to utilize C-Ent to do HPO but only evaluated it on their own adaptation method. 
We evaluate C-Ent extensively, across various adaptation methods, datasets, and vision tasks. We reveal that C-Ent is very effective for HPO for several adaptation methods, but also expose a critical issue in this approach, which is that it cannot detect the collapse of  neighborhood structure in target samples (See Fig.~\ref{fig:idea}). 
By \textit{neighborhood structure}, we mean the relationship between samples in a feature space. 
Before any adaptation, target samples embedded nearby are highly likely in the same class (\textit{No adaptation} in Fig.~\ref{fig:idea}). A good UDA model will keep or even enhance the relationships of target samples while aligning them to the source.
However, a UDA model can falsely align target samples with the source and incorrectly change the neighborhood structure ((\textit{DANN} in Fig.~\ref{fig:idea}). 
Even in this case, C-Ent can become small and choose a poorly-adapted model. 

Natekar \etal~\cite{natekar2020representation} measure the consistency of feature embeddings within a class and their discrepancy from other classes to predict the generalization of supervised models by using labeled samples. 
Accounting for such relationships between points is a promising way to overcome the issues of C-Ent. But, since computing these metrics requires labeled samples, we cannot directly apply this method. 

This leads us to propose a novel unsupervised validation criterion that considers the neighborhood density of the unlabeled target. Our notion of a neighborhood is soft, i.e. we do not form explicit clusters as part of our metric computation. Rather, we define soft neighborhoods of a point using the distribution of its similarity to other points, and measure density as the entropy of this distribution. We assume that a well-trained source model will embed target samples of the same class nearby and thus will form dense implicit neighborhoods. 
The consistency of representations within each neighborhood should be preserved or even enhanced by a well-adapted model.
Monitoring the density thus enables us to detect whether a model causes the collapse of implicit neighborhood structure as shown in Fig.~\ref{fig:idea}.

Our proposed metric, called \textit{\ours (\oursv)}, is simple yet more effective than competing validation methods. Rather than focusing on the source and target relationship (like IWV~\cite{sugiyama2007covariate} or DEV~\cite{you2019towards}), we measure the discriminability of target features by computing neighborhood density and choosing a model that maximizes it. 

We demonstrate that target accuracy is consistent with our criterion in many cases. Empirically, we observe that \oursv works well for closed and partial domain adaptive image classification, as well as for domain adaptive semantic segmentation. \oursv is even effective in choosing a suitable source domain given an unlabeled target domain.

Our contributions are summarized as follows:

\begin{itemize}
     \vspace{-3mm}
    \item We re-evaluate existing criteria  for UDA and call for more practical validation of DA approaches. 
     \vspace{-3mm}
    \item We propose \ours metric which considers target neighborhood structure, improves upon class entropy (C-Ent) and achieves performance close to optimal (supervised) HPO in 80\% cases on closed, partial DA, and domain adaptive semantic segmentation.  
\end{itemize}

\begin{table}[t]
\begin{center}
    \scalebox{0.8}{
\begin{tabular}{l|ccc|c}
	\toprule[1.0pt]
\multirow{2}{*}{Method}  & \multicolumn{3}{c|}{Technical Advantages}&Stability\\\cline{2-4}
&  w/ $X_t$&  w/o $X_s, Y_s$& w/o HP &across methods\\\hline
\bf{Source Risk} &\xmark&\xmark&Test split&\xmark\\
\bf{IWV}~\cite{sugiyama2007covariate,you2019towards}&\xmark&\xmark&+Density Model&\xmark\\
\rowcolor{Gray}
\bf{Entropy}~\cite{morerio2017minimal}&\checkmark&\checkmark&\checkmark&\xmark\\
\rowcolor{Gray}
\bf{\oursv(Ours)} &\checkmark&\checkmark&\checkmark&\checkmark\\
	\bottomrule[1.0pt]
\end{tabular}}
\end{center}
\vspace{-6mm}
\caption{{\small Technical comparison with other validation approaches. $X_t$ denotes unlabeled target, and ($X_s, Y_s$) denote labeled source samples. \oursv computes a score on unlabeled target samples. Empirically, we verify that our method is stable across different datasets, methods, and tasks. }}
\vspace{-15pt}
\label{tb:compare_method}

\end{table}

 \vspace{-5mm}
\section{Related Work}
\textbf{Domain Adaptation} aims to transfer knowledge from a labeled source domain to a label-scarce target domain. 
Its application in vision is diverse: image classification, semantic segmentation~\cite{hoffman2016fcns, tsai2018learning}, and object detection~\cite{dafaster}. 
One popular approach in DA is based on distribution matching by adversarial learning~\cite{ganin2014unsupervised, tzeng2017adversarial, long2017conditional}. Adversarial adaptation seeks to minimize an approximate domain discrepancy distance through an adversarial objective with respect to a domain discriminator. Recently, some techniques that utilize clustering or a variant of entropy minimization have been developed~\cite{saito2020universal,kang2019contrastive, tang2020unsupervised, jin2020minimum}. \cite{tang2020unsupervised, kang2019contrastive} propose to train a model by minimizing inter-class discrepancy given the number of classes. \oursv computes the density of local neighborhoods and selects a model with the largest density, which allows us to select a good model without knowing the number of classes in the target. 
All of the existing approaches have important hyper-parameters to be tuned, such as the trade-off parameter between source classification and adaptation loss. Another important hyper-parameter is softmax temperature~\cite{saito2020universal,jin2020minimum}. 
 Given a specific target domain, the selection of a source domain is also important. If we have to choose only a single source domain, the selection process is crucial for the model's performance. However, it is questionable whether existing approaches do HPO in a realistic way.

\textbf{Validation Methods for UDA.}
In Table~\ref{tb:compare_method}, we summarize several prior validation methods that do not need any target labeled samples. The validation methods themselves can have hyper-parameters (HP) and other requirements. 

\noindent\textbf{Source Risk.}
Ganin \etal~\cite{ganin2014unsupervised} considers the source risk to select hyper-parameters. But, the source risk is not a good estimator of the target risk in the presence of a large domain gap.

\noindent\textbf{Importance Weighted Validation (IWV) and  DEV.} Sugiyama \etal~\cite{sugiyama2007covariate} and You \etal~\cite{you2019towards} validate methods using the risk of source samples. If a source sample is very similar to target samples, the risk on the source sample is heavily weighted. This approach has a similar issue to Source Risk validation. Since DEV is developed to control the variance in IWV, we use it as a baseline in experiments.

\noindent\textbf{Entropy (C-Ent).} 
Morerio~\etal~\cite{morerio2017minimal} employ entropy of classification output. If the model has confident predictions in classifying target samples, the hyper-parameter is considered appropriate. The method is simple and does not require validation with labeled samples. But, Morerio \etal~\cite{morerio2017minimal} apply the C-Ent criterion only to tune their proposed model, which makes its applicability to diverse methods unclear. We extensively evaluate this approach and reveal that while it is often useful, it has a critical failure mode. 
This failure mode is that domain adaptation models can output confidently incorrect predictions for target samples.
In comparison, we empirically show that \ours gives the most stable results across different datasets and methods for both image classification and semantic segmentation.  

\textbf{Locally Linear Embedding (LLE).} 
Roweis \etal~\cite{roweis2000nonlinear} compute low dimensional and neighborhood-preserving embeddings of high dimensional data. The neighborhood-preserving embeddings recover global nonlinear structure. We aim to pick a model by monitoring the density of implicit neighborhoods during adaptation.

\begin{figure*}
    \centering
    \includegraphics[width=\linewidth]{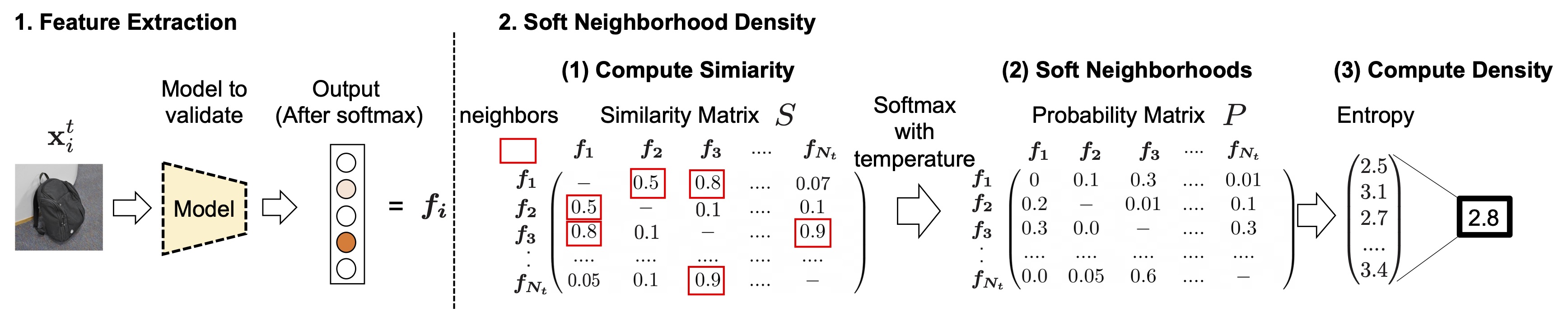}  
    \vspace{-8mm}
    \caption{{\small {\bf Method overview.} \ours measures the density of implicit local neighborhoods in the target domain, which is used to select hyper-parameters of the adaptation model. We first extract features from the softmax layer for all target samples. We then compute the similarity distribution from the pair-wise similarities of the features (red boxes highlight similar points). Finally we use the entropy of the similarity distribution as our evaluation criterion (\oursv), where the higher its value, the better.}}
    \label{fig:pipeline}
    \vspace{-15pt}
\end{figure*}
\section{Approach}
\textbf{Problem Setting.} In UDA, we are provided labeled source data $\mathcal{D}_{s}=\left\{\left(\mathbf{x}_{i}^{s}, {y_{i}}^{s}\right)\right\}_{i=1}^{N_{s}}$ and unlabeled target data $\mathcal{D}_{t} = \left\{\left(\mathbf{x}_{i}^{t} \right)\right\}_{i=1}^{N_{t}}$. 
Generally, domain adaptation methods optimize the following loss,
 \begin{equation}
L = L_{s}(x_s, y_s) + \lambda L_{adapt}(x_s, x_t, \eta),\label{eq:adapt}
 \end{equation}
where $L_{s}$ is the classification loss for source samples and $L_{adapt}$ is the adaptation loss computed on target samples. 
$\lambda$ controls the trade-off between source classification and adaptation loss. $\eta$ are hyper-parameters used to compute the adaptation loss. Our goal is to build a criterion that can tune $\lambda$ and $\eta$. 
We additionally aim to choose the best training iteration since a model can be sensitive to its choice. 

\textbf{Assumptions.}
We assume that target samples embedded nearby should belong to the same class. Therefore, representations of a well-adapted model should result in dense neighborhoods of highly-similar points. 
We express density as the consistency of representations within each soft neighborhood, such that \ours becomes large with a well-adapted model. 
Models are trained with labeled source samples as well as unlabeled target samples. Since source and target are related domains, the model will have somewhat discriminative features for target samples. Before adaptation, such features will define \textit{initial neighborhoods}, namely, samples with higher similarity to each other relative to the rest of the points. If the selection of a method and hyper-parameter is appropriate, such neighborhoods should be preserved rather than split into smaller clusters, and the similarity of features within soft neighborhoods should increase.

\textbf{Motivation.}
First, we empirically show the necessity of considering the neighborhood structure. 
Class Entropy (C-Ent)~\cite{morerio2017minimal} measures the confidence in prediction.
Through domain-adaptive training with inappropriate hyper-parameters, the confidence in the target prediction can increase while drastically changing the neighborhoods from the initial neighborhood structure.

A toy example (Fig.~\ref{fig:idea}) clarifies this idea.
We generate source data from two Gaussian distributions with different means, which we regard as two classes. Then, we obtain target data by shifting the mean of one of the Gaussians. 
We train two-layered neural networks in two ways: training with distribution alignment (DANN~\cite{ganin2014unsupervised}, $\lambda=1.0$) and training with only source samples ($\lambda=0$). From the input space, the target samples should not be aligned with source class 1 (green circles). But, the DANN model aligns target samples incorrectly and confidently misclassifies many target samples, resulting in a very small C-Ent. If we employ C-Ent as a criterion, we will select the left model which performs poorly. But, if we are able to consider whether the density of neighborhoods is maintained or increased, we can avoid selecting the poor model.

In this toy example, we have utilized the partial DA setting~\cite{cao2018partial}, where the target label set is the subset of the source, and DANN~\cite{ganin2014unsupervised} to illustrate the issue of C-Ent.
In fact, this kind of \textit{wrong alignment} can happen in real data when we use distribution alignment (Fig.~\ref{fig:feat_entropy}). 

\subsection{\ours (\oursv)}\label{sec:wind}
 \vspace{-2mm}
 We aim to design a criterion that can consider the density of implicit local neighborhoods in the target domain. 
 We define a soft \textit{neighborhood} for each sample by computing its similarity to other samples in the dataset, and converting it to a probability distribution. This is done via a softmax activation with temperature-scaling to ignore samples outside of the local neighborhood. Once we define the soft neighborhoods, we can estimate their density by computing the entropy of the distribution. The overall pipeline in Fig.~\ref{fig:pipeline} consists of 1) computing the similarity between samples, 2) applying the softmax activation with temperature-scaling (identifying soft neighborhoods), and 3) calculating soft neighborhood density. 
 
\textbf{Similarity Computation.}
 We first compute similarity between target samples, $S \in \mathbb{R}^{{N_t}\times{N_t}}$, where $N_t$ denotes the number of target samples. 
Let $S_{ij} = \sprod{\vec{f}_i^t}{\vec{f}_j^t}$, where $\vec{f}_i^t$ denotes a $L_2$ normalized target feature for input $\mathbf{x}_i^{t}$. We ignore diagonal elements of $S$ because our goal is to compute the distances to neighbors for each sample. 
This matrix defines distances, but it is not clear which samples are relatively close to each other vs. far away just from this matrix. 

\textbf{Soft Neighborhoods.}
To highlight the difference between nearby points and far-away points, we convert the similarity between samples into a probabilistic distribution, $P$, using temperature scaling and the softmax function,  
\begin{equation}\label{eq:distribution}
    P_{ij} = \frac{\exp(S_{ij}/\tau)}{\sum_{j'}\exp(S_{ij'}/\tau)} \;, 
     \vspace{-2mm}
\end{equation}
where $\tau$ is a temperature constant.
Note that the temperature scaling and the exponential function have the ability to enlarge the difference between similarity values $S_{ij}$. Therefore, if a sample $j$ is relatively dissimilar from $i$, the value of $P_{ij}$ will be very small, which allows us to ignore distant samples for the sample $i$. The temperature is the key to implicitly identifying neighborhoods; we set it to 0.05 across all experiments given the results of the toy dataset. 

\textbf{\ours.}
Next, we design a metric to consider the neighborhood density given $P$. The metric needs to evaluate the consistency of representations within the implicit neighborhoods. If a model extracts ideally discriminative features, the representations within a neighborhood are identical. To identify this situation, we propose to compute the entropy of $P$. For example, if the entropy of $P_{i}$ is large, the probabilistic distribution should be uniform across the soft neighborhoods; that is, neighbors of the sample $i$ are concentrated into very similar points. 
Specifically, we compute the entropy for each sample and take the average of all samples as our final metric: 
\vspace{-3mm}
\begin{equation}\label{eq:entropy}
 H(P) = -\frac{1}{N_t}\sum_{i=1}^{N_t}\sum_{j=1}^{N_t} P_{ij}  \log P_{ij}. 
 \end{equation}
 \vspace{-2mm}

We then choose the model that has the highest entropy of all candidate models. If a model falsely separates samples into clusters as in Fig.~\ref{fig:idea}, the entropy becomes small.

 \textbf{Input Features.} 
 The key to the success of our method is extracting implicit neighborhoods. Ideally, all samples in the same class should be embedded near each other. In that case, Eq.~\ref{eq:entropy} can compute the density of each class. 
 Hence, we need to use class-discriminative features, and the choice of features can be essential. We propose to employ the classification softmax output as our input feature vector $\vec{f}$. Since this feature represents class-specific information, it is the most likely to place target samples of the same class together.
As we analyze in experiments, this feature has smaller within-class variance than either middle-layer features or classification output without softmax activation. 
Therefore, our hope is that the number of clusters in Eq.~\ref{eq:distribution} should be equal to (closed set case) or less than (partial set case) the number of source classes. 
In the ideal case, the computed density should be close to the within-class density.

 \textbf{Discussion.}
Note that we assume that the classifier is well-trained on source samples. The criterion becomes very large if a model produces the same output for all target samples. To avoid this, the model needs to be well trained on source samples, which can be easily monitored by seeing the loss on source training samples. 

Also, note that our assumption is that samples embedded nearby are likely to share the category labels. This is consistent with an assumption made by general domain adaptation methods~\cite{ben2010theory}. If a pre-trained model provides very poor representations or the source domain is too different from the target, the assumption will not be satisfied. Under this setting, any metrics of UDA will not be a good tool to monitor the training. 

\textbf{Extension to Semantic Segmentation.}\quad 
In semantic segmentation, the outputs for each image can be as large as one million pixels. When the number of target samples is that large, the computation of the similarity matrix can be very expensive. In order to make the computation of \oursv more efficient, we subsample the target samples to make the similarity graph smaller and easier to compute. In our experiments, we randomly sample a hundred pixels for each image and compute \oursv for each, then, take the average of all target images. Although the method is straightforward, our experiments show that the resulting approximation of \oursv continues to be an effective criterion for selecting hyper-parameters.

\begin{table*}[t]
\scalebox{0.82}{
\begin{tabular}{l|cccc|cccc|cccc|cccc||c}
	\toprule[1.0pt]
	\multirow{2}{*}{Method}  &  \multicolumn{4}{c|}{CDAN~\cite{long2017conditional}} &\multicolumn{4}{c|}{MCC~\cite{jin2020minimum}}  &\multicolumn{4}{c|}{NC~\cite{saito2020universal}}&\multicolumn{4}{c||}{PL~\cite{lee2013pseudo}}&\multirow{2}{*}{Avg}     \\
 & A2D & W2A & R2A& A2P &A2D & W2A & R2A& A2P &A2D & W2A & R2A& A2P&A2D & W2A & R2A& A2P&\\\hline

 \bf{Lower Bound} & 77.7 & 57.4 & 58.7 & 59.9 & 80.4 & 62.6 & 59.0 & 60.5 & 69.8 & 60.2 & 67.7 & 66.5 & 80.4 & 65.1 & 66.8 & 66.6 & 66.4  \\\hline 
\bf{Source Risk} & 87.9 & 65.5 & 62.3 & 62.2 & \bf{92.8} & 70.2 & 65.6 & 72.6 & 78.1 & \bf{66.8} & 71.3 & 70.4 & 84.5 & 67.4 & 69.3 & 67.4 & 71.1  \\
\bf{DEV~\cite{you2019towards}}        & 90.0 & 66.4 & 63.5 & 62.4 & 91.3 & 67.6 & 63.1 & 70.1 & 78.1 & 65.3 & 71.4 & 72.1 & 84.7 & \bf{67.5} & 69.0 & 69.2 & 71.6 \\
\rowcolor{Gray}
\bf{Entropy~\cite{morerio2017minimal}}     & 82.3 & 63.8 & 61.7 & 63.4 & 91.3 & \bf{72.4} & 66.9 & 70.3 & \bf{88.4} & 66.4 & 72.0 & 73.7 & 84.8 & 67.4 & \bf{70.1} & \bf{69.4} & 71.9 \\
\rowcolor{Gray}
\bf{\oursv(Ours)}        & \bf{92.9} & \bf{67.0} & \bf{70.8} & \bf{67.3} & 92.6 & 67.4 & \bf{68.8} & \bf{72.8} & \bf{88.4} & 66.4 & \bf{72.5} & \bf{73.9} & \bf{85.1} & 67.4 & \bf{70.1} & \bf{69.4} & \bf{74.3}  \\\hline
\bf{Upper Bound} & 93.3 & 69.8 & 71.1 & 68.1 & 94.5 & 72.9 & 69.5 & 74.4 & 89.6 & 71.2 & 72.8 & 74.1 & 86.8 & 68.1 & 70.8 & 70.2 & 75.5  \\
\bottomrule[1.0pt]
\end{tabular}}
\vspace{-3mm}
\caption{{\small \textbf{Results of closed DA.} \oursv provides faithful results for all methods and datasets, \ie Office (A2D and W2A) and OfficeHome (R2A and A2P) whereas baselines show several failure cases. Lower/Upper bounds are results obtained with the worst/best model.}}
\label{tb:closed}
 
\end{table*}

\begin{table*}[]
 \vspace{-3mm}
\begin{center}
\scalebox{0.9}{
    \begin{tabular}{l|ccc|ccc|ccc|ccc||c}
	\toprule[1.0pt]
	\multirow{2}{*}{Method}  &   \multicolumn{3}{c|}{CDAN~\cite{long2017conditional}} &\multicolumn{3}{c|}{MCC~\cite{jin2020minimum}}  &\multicolumn{3}{c|}{NC~\cite{saito2020universal}}&\multicolumn{3}{c||}{PL~\cite{lee2013pseudo}}&\multirow{2}{*}{Avg}      \\
 & R2A&P2C &A2P&R2A&P2C &A2P &R2A&P2C &A2P& R2A&P2C&A2P&\\\hline

\bf{Lower Bound} & 60.9 & 34.5 & 60.2 & 53.7  & 38.3  & 60.9  &60.0 & 37.5 & 52.3  & 49.2 & 56.6 & 44.1 & 50.7 \\\hline
\bf{Source Risk}& \bf{67.6} & 42.0    & 64.8& 65.1 & 47.4 & 73.3 & 76.6 & 49.8 & 77.9  & 61.5 & 72.7 & 53.5 & 62.7 \\
\bf{DEV~\cite{you2019towards}}&65.6 & 36.6  & 63.9 & 67.7 & 47.3  & 70.3  & 72.3 & 54.5 & 67.0  & 62.4 & 66.3 & 52.1 & 60.5 \\
\rowcolor{Gray}
\bf{Entropy~\cite{morerio2017minimal}}&64.7 & 40.3  & 64.8& 53.8 & 40.4  & 62.0 &  79.1 & \bf{58.2} & \bf{78.7}  & 60.1 & 71.7 & 47.0 & 60.1 \\
\rowcolor{Gray}
\bf{\oursv(Ours)}&66.3 & \bf{45.7} & \bf{65.4}& \bf{70.2} & \bf{50.8}  & \bf{79.3}  &  \bf{79.2} & 58.1 & 78.2   & \bf{68.7} & \bf{72.2} & \bf{57.9} & \bf{66.0} \\\hline
Upper Bound &68.8 & 46.9  & 68.5 & 72.0 & 52.1 & 79.7  & 80.1 & 58.2 & 79.1 
 & 69.0 & 74.0 & 59.4 & 67.3\\

	\bottomrule[1.0pt]
\end{tabular}}
\vspace{-3mm}
\caption{{\small \textbf{Results of partial DA on OfficeHome.} \oursv performs the best on average. }}
\label{tb:partial}\vspace{-20pt}

\end{center}

\end{table*}

\vspace{-3mm}
\section{Experiments}
 \vspace{-2mm}
First, we evaluate the existing metrics and \oursv to choose hyper-parameters in domain adaptation for image classification and semantic segmentation. 
Second, we show experiments to analyze the characteristics of our method. 

We evaluate the ability to choose suitable hyper-parameters including checkpoints (\ie, training iterations) for unsupervised domain adaptation. Closed DA (CDA) assumes that the source and target domain share the same label set while partial DA (PDA) assumes that the target label set is the subset of the source. We perform experiments on both adaptation scenarios.
The details of semantic segmentation are described in the appendix. The general design of the experiment is similar to image classification. 

\vspace{-2mm}
\subsection{Adaptation Methods}
\vspace{-2mm}
For each method, we select the hyper-parameter mentioned below, plus a training iteration. See appendix for other hyper-parameters used in experiments.

\textbf{Adversarial Alignment.}\quad As a representative method of domain alignment, we use CDAN~\cite{long2017conditional}. We select a weight of the trade-off between domain confusion loss and source loss ($\lambda$ = 0.1, 0.3, 0.5, 1.0 ,1.5, with $\lambda$ = 1.0 as
its default setting). 
To analyze the behavior in detail, we also utilize DANN~\cite{ganin2014unsupervised} for OfficeHome PDA. The validation is done in the same way as CDAN.

\textbf{Clustering.}\quad As a clustering-based DA method, we utilize Neighborhood Clustering (NC)~\cite{saito2020universal}. NC performs clustering using similarity between features and a temperature is used to compute the similarity distribution. 
Since the selection of the temperature value can affect the performance, we evaluate criteria to choose the best temperature ($\eta$ = 0.5, 0.8, 1.0, 1.5, with $\eta$ = 1.0 as its default setting).

\textbf{Classifier Confusion.}\quad As a recent state-of-the art method, we employ MCC~\cite{jin2020minimum}, where a temperature is used to compute the classifier's confusion loss. The goal is to tune the temperature values ($\eta$ = 1.5, 2.0, 2.5, 3.0, 3.5, with $\eta$ = 2.5 as its default setting). 

\textbf{Pseudo Labeling (PL).}\quad Employing pseudo labels~\cite{lee2013pseudo} is one of the popular approaches in DA~\cite{zou2019confidence,zou2018unsupervised,saito2017asymmetric}. 
One important hyper-parameter is a threshold to select confident target samples. If the output probability for a predicted class is larger than the threshold, we use the sample for training. The optimal threshold may be different for different datasets. Then, our goal is to tune the threshold values ($\eta$ = 0.5, 0.7, 0.8, 0.9, 0.95). 

\textbf{Semantic Segmentation.}\quad AdaptSeg~\cite{tsai2018learning} and ADVENT~\cite{vu2018advent} are employed. For both methods, the goal is to tune both trade-off parameters of adversarial alignment loss ($\lambda$) and training iterations.

\subsection{Setup}
 \vspace{-2mm}
\textbf{Datasets.}\quad
For image classification, we use Office~\cite{saenko2010} (Amazon to DSLR (A2D) and Webcam to Amazon (W2A) adaptation) with 31 categories and OfficeHome~\cite{venkateswara2017Deep} (Real to Art (R2A), Art to Product (A2P) and Product to Clipart (P2C)) with 65 categories. Office is used for CDA while we use OfficeHome for both CDA and PDA. For the category split of PDA, we follow ~\cite{ETN_2019_CVPR}. To further demonstrate applicability to large-scale datasets, we evaluate \oursv on VisDA~\cite{peng2017visda} and DomainNet~\cite{peng2018moment} in CDA. We describe the detail in the appendix.
In semantic segmentation, we use GTA5~\cite{Richter_2016_ECCV} as a source and Cityscape~\cite{cordts2016cityscapes} as a target domain. 

\textbf{Baselines.}\quad
Entropy~\cite{morerio2017minimal} directly employs the entropy of the classification output. It takes the average of all samples. A smaller value should indicate a better-adapted model. For DEV~\cite{you2019towards}, we need to have held-out validation source samples. Since holding out many source samples can degrade the accuracy of adapted models, we take 3 source samples per class as validation sets. Increasing the number of source validation samples to more than 3 per class does not much improve the validation performance. See the appendix for more detail.
A smaller risk represents a better model. Similarly, Source Risk is measured on the source validation samples. We also report lower bound and upper bound performance among all hyper-parameters. 

\textbf{Evaluation Protocol.}\quad
In image classification, we train all adaptation methods for 10,000 iterations. Although every method has different default training iterations, we keep them the same for the simplicity of experiments. Then, we select a checkpoint that shows the best value among all reported iterations and choice of hyper-parameters. In semantic segmentation, we calculate mIoU and each criterion similarly. We don't use DEV~\cite{you2019towards} for semantic segmentation since the design of the domain classifier is complicated. We run experiments three times and show their averaged accuracy. 

\textbf{Implementation.}\quad
We utilize published official implementations for adaptation methods, CDAN, MCC, NC, AdaptSeg, and ADVENT. For the pseudo labeling method, we use the NC's implementation. These methods use ResNet50 or ResNet101~\cite{he2016deep} as backbone networks. AdaptSeg and ADVENT employ DeepLab~\cite{chen2017deeplab}. See the appendix for more details.

\begin{figure}[!t]
\centering
\begin{subfigure}[t]{3in}
\centering
\includegraphics[width=\linewidth]{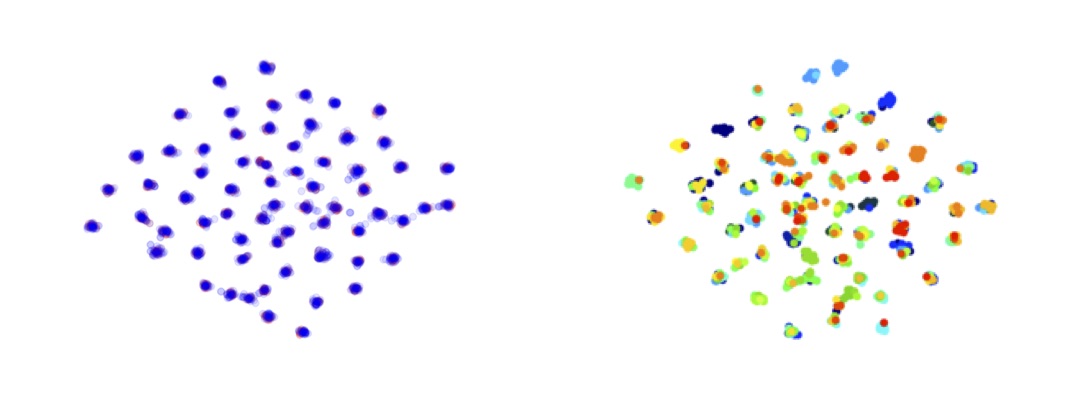}
\vspace{-8mm}
\caption{DANN tuned by Entropy~\cite{morerio2017minimal} (Accuracy: 35.1 \%).}
\label{fig:feat_entropy}
\end{subfigure}
\begin{subfigure}[t]{3in}
 \centering
\includegraphics[width=\linewidth]{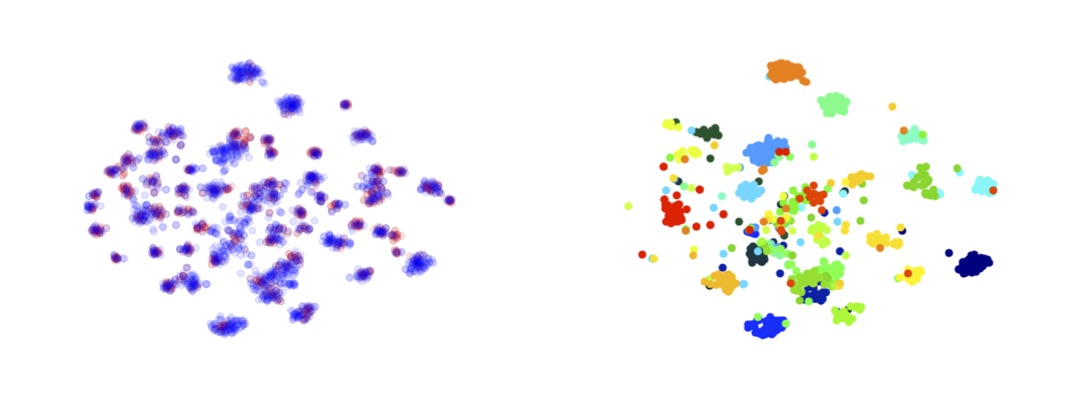}
\vspace{-8mm}
\caption{DANN tuned by \oursv (Accuracy: 70.8 \%).}
\label{fig:feat_ours}
\end{subfigure}
\vspace{-3mm}
\caption{{\small \textbf{Feature visualization~\cite{maaten2008visualizing}.} OfficeHome partial DA using DANN~\cite{ganin2014unsupervised}. \textbf{Left:} Source (Blue), Target (Red). \textbf{Right:} Target samples. Different colors denote different classes. (a) Entropy~\cite{morerio2017minimal} does not detect the collapse of the neighborhood structure and chooses a model that wrongly aligns features. (b) \oursv chooses a model that keeps the structure.}}
\vspace{-15pt}
\label{fig:features}
\end{figure}
\begin{table*}[]
\begin{center}
\scalebox{0.9}{

\begin{tabular}{l|cc|cc|cc|cc||c}
	\toprule[1.0pt]
	\multirow{2}{*}{Method}  &  \multicolumn{2}{c|}{CDAN~\cite{long2017conditional}} &\multicolumn{2}{c|}{MCC~\cite{jin2020minimum}}  &\multicolumn{2}{c|}{NC~\cite{saito2020universal}}&\multicolumn{2}{c||}{PL~\cite{lee2013pseudo}} & \multirow{2}{*}{Avg}\\
 & VisDA &DNet &VisDA &DNet &VisDA &DNet&VisDA &DNet \\\hline
 
\bf{Lower bound} & 51.1$\pm$1.3& 51.3$\pm$1.4& 67.1$\pm$1.1& 54.8$\pm$2.1& 44.8$\pm$3.7& 56.9$\pm$0.4& 58.7$\pm$1.1& 55.5$\pm$0.3& 55.0$\pm$1.1\\\hline
\bf{Source Risk} & \bf{72.6$\pm$0.8} & 63.8$\pm$1.4& 71.7$\pm$0.8& 58.7$\pm$0.8& 65.8$\pm$1.3 & 62.0$\pm$0.2 & 66.7$\pm$2.8 & 59.9$\pm$0.5& 65.1$\pm$0.3\\
\bf{DEV~\cite{you2019towards}}         & \bf{72.6$\pm$0.8}& 57.9$\pm$4.1 & 72.3$\pm$3.0 & 58.5$\pm$0.5& 65.8$\pm$1.3 & 59.4$\pm$0.8& 66.7$\pm$2.8& 59.1$\pm$0.3& 64.0$\pm$1.1 \\
\rowcolor{Gray}
\bf{Entropy~\cite{morerio2017minimal}}& 69.9$\pm$1.8 & 61.5$\pm$1.1 & 68.9$\pm$1.3 & \bf{59.2$\pm$0.2} & \bf{68.4$\pm$1.1} & 62.3$\pm$0.6 & 68.5$\pm$0.1 & 60.7$\pm$0.3 & 64.9$\pm$0.2 \\
\rowcolor{Gray}
\bf{\oursv(Ours)}      & 70.3$\pm$0.1 & \bf{64.9$\pm$0.4} & \bf{73.0$\pm$1.1} & 58.9$\pm$2.3 & 66.9$\pm$3.2 & \bf{62.4$\pm$0.8} & \bf{69.0$\pm$1.1} & \bf{60.9$\pm$0.1} & \bf{65.8$\pm$0.3} \\\hline
\bf{Upper bound} & 74.1$\pm$1.1 & 65.6$\pm$0.1 & 74.5$\pm$0.6 & 61.2$\pm$0.5 & 69.2$\pm$0.2 & 63.2$\pm$0.2 & 69.2$\pm$0.8 & 61.0$\pm$0.1 & 67.2$\pm$0.1\\
\bottomrule[1.0pt]
\end{tabular}}
\vspace{-2mm}
\caption{{\small \textbf{VisDA~\cite{peng2017visda} and DomainNet (DNet)~\cite{peng2018moment} results in closed DA.} Averaged accuracy over three runs and its standard deviation are shown. We utilize Real to Clipart adaptation for DomainNet. \oursv performs the best on average. }}
\label{tb:dnet}
\vspace{-2mm}

\end{center}
\end{table*}

\begin{figure*}
\vspace{-6mm}
    \centering
    \includegraphics[width=\linewidth]{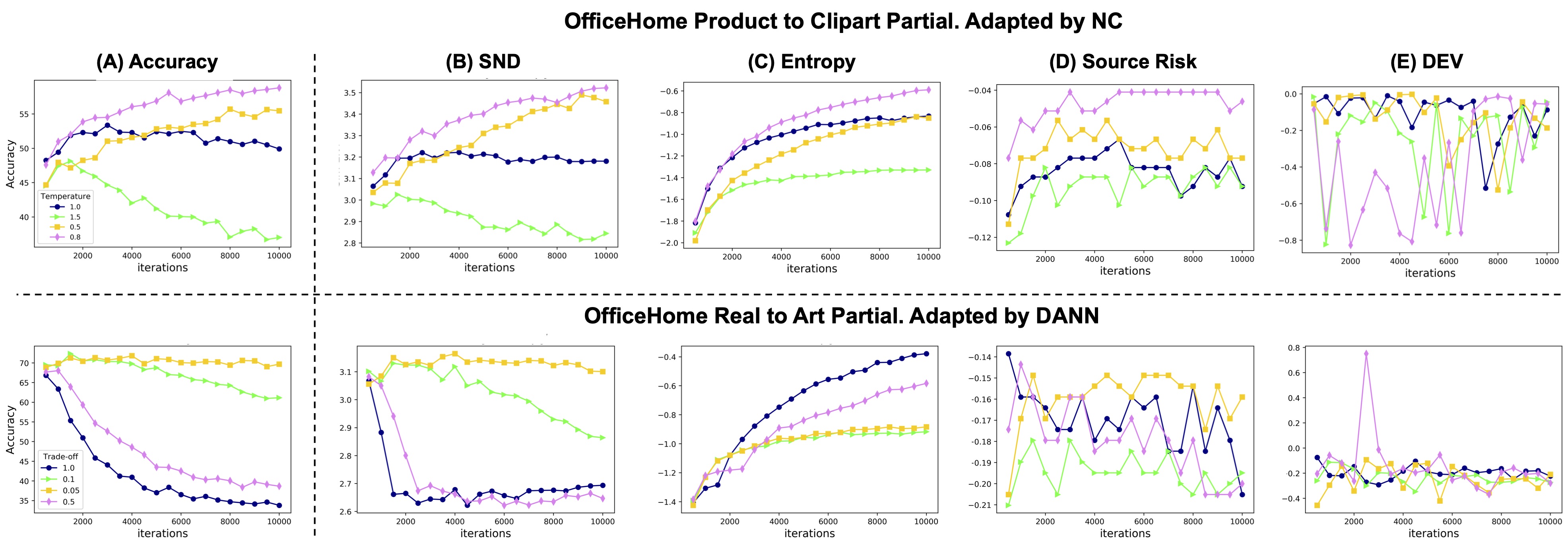}
    \vspace{-7mm}
    \caption{{\small Iteration versus accuracy and HPO criteria. To ease comparison between accuracy and criteria, we flip the sign of criteria for Entropy, Source risk, and DEV. \textbf{Notice how  \oursv curves track the Accuracy. More results are shown in appendix.}}}
    \label{fig:graph_summary}
    \vspace{-10pt}
\end{figure*}
\begin{table}[]
\small
\centering
\begin{tabular}{l|c|c}
	\toprule[1.0pt]
GTA5 to Cityscape&AdaptSeg~\cite{tsai2018learning}  & ADENT~\cite{vu2018advent} \\\hline
\bf{Lower Bound}&31.1$\pm$0.5&21.0$\pm$2.5\\\hline 
\bf{Source Risk} &35.6$\pm$2.9&37.2$\pm$1.6\\
\rowcolor{Gray}
\bf{Entropy~\cite{morerio2017minimal}} &\bf{39.3$\pm$1.1}&35.6$\pm$3.7\\
\rowcolor{Gray}
\bf{\oursv(Ours)} &\bf{39.5$\pm$0.8}&\bf{40.2$\pm$0.5}\\ \hline 
\bf{Upper Bound}&40.4$\pm$0.5&43.6$\pm$0.3 \\ 
	\bottomrule[1.0pt]
\end{tabular}
\vspace{-3mm}
\caption{{\small Validation results in domain-adaptive semantic segmentation in GTA5 to Cityscapes adaptation.}}
\label{tb:seg}
\vspace{-15pt}
\end{table}

 \vspace{-2mm}
\subsection{Validation Results}
\textbf{Image Classification.}\quad
The results of image classification are summarized in Tables~\ref{tb:closed}, \ref{tb:partial}, and~\ref{tb:dnet}. Fig.~\ref{fig:graph_summary} shows plots of accuracy and criteria for several adaptation settings. In most cases, DA methods are sensitive to hyper-parameters and training iterations. 
As we can see in the Tables, our proposed method faithfully selects good checkpoints for various methods and two category shifts. Of course, there are some gaps between the upper bound and our score, but the gap is not large. Importantly, \oursv does not catastrophically fail in these cases while other criteria choose several bad checkpoints. Besides, the curve of accuracy and \oursv have very similar shapes. The results denote that \oursv is effective for HPO in various methods. In the experiments on VisDA and DomainNet, most methods select good checkpoints since adaptation methods are stable across different hyper-parameters.

\textit{Source Risk} provides a good model in some cases, but also catastrophically fails in some cases such as A2D in NC Table~\ref{tb:closed}. 
DEV~\cite{you2019towards} also sometimes catastrophically fails. From Fig.~\ref{fig:graph_summary}, these two criteria have some variance and do not necessarily reflect the accuracy in the target. We hypothesize that there are two reasons. First, Source Risk is not necessarily correlated with the performance of the target. If a model focuses on the classification of source samples (\ie, setting $\lambda=0$ in Eq. \ref{eq:adapt}), the risk gets small, which does not indicate good performance on the target domain. Unless the source and target are very similar, the risk will not be reliable. Second, we may need careful design in selecting validation source samples and the domain classifier construction for DEV. However, considering the practical application, the validation methods should not have a module that requires careful design. 

\textit{Entropy}~\cite{morerio2017minimal} (C-Ent) shows comparable performance to \oursv in NC~\cite{saito2020universal} and PL~\cite{lee2013pseudo}. This is probably because both methods are trained to maintain the target neighborhood structure. Then, by monitoring the confidence of the prediction, we can pick a good model.
However, as is clear from the graph at the bottom of Fig.~\ref{fig:graph_summary}, it causes catastrophic failure by mistakenly providing confident predictions in PDA (Real to Art) adapted by DANN. 
We show the feature visualization of DANN results in Fig.~\ref{fig:features}. The model selected by C-Ent collapses the neighborhood structure of target samples yet matches them with source samples, which results in a lower C-Ent value. By contrast, \oursv selected a model that maintains the structure. Since the overconfidence issue can happen in many methods and datasets, C-Ent is not reliable for some methods. This is consistent with the results on the toy dataset in Fig.~\ref{fig:idea}.

\textbf{Semantic Segmentation.}
Table~\ref{tb:seg} describes the checkpoint selection result. \oursv selects good checkpoints for both methods. 
If we compare the performance of the upper bound, ADVENT~\cite{vu2018advent} is better than AdaptSeg~\cite{tsai2018learning} with 3.2 points in mIoU. But, the gap becomes much smaller (only 0.7 points) if we apply unsupervised evaluation. ADVENT~\cite{vu2018advent} is more sensitive to hyper-parameters such as training iterations than AdaptSeg~\cite{tsai2018learning}. 
Many current state-of-the-art models seem to select checkpoints by the target risk. But, as this result indicates, such comparisons may be misleading for real-world applications. 
\begin{table}[t]
\begin{center}
\scalebox{0.85}{
    \begin{tabular}{l|cc|cc|cc|cc}
	\toprule[1.0pt]
	\multirow{2}{*}{Method}  &   \multicolumn{2}{c|}{Target: R}& \multicolumn{2}{c|}{Target: Ar} & \multicolumn{2}{c|}{Target: Cl}  & \multicolumn{2}{c}{Target: Pr}\\
 & Acc. & S&Acc. & S&Acc. & S&Acc. & S\\\hline
\bf{\small Lower bound}&67.5&Cl&56.3&Pro&43.2&Pro&64.4&Cl\\\hline 
\rowcolor{Gray}
\bf{\oursv(Ours)}&74.6&Ar&69.2&R&50.1&R&78.3&R\\\hline 
\bf{\small Upper bound}&76.0&Pro&69.6&R&51.1&R&79.0&R\\
\bottomrule[1.0pt]
\end{tabular}}
\label{label 2}
\end{center}
\vspace{-7mm}
\caption{{\small Source domain selection experiments using the Office-Home dataset. 
We show the accuracy of the selected model and selected source domain (R: Real, Ar: Art, Cl: Clipart, Pr: Product). \oursv selects the best source domain except for the Real domain. Even in that case, the selected model and the oracle accuracy perform similarly.} }
\label{tb:source_select}
\vspace{-3mm}

\end{table}

\begin{table}[t]
\begin{center}
\scalebox{0.85}{
    \begin{tabular}{l|ccc}
	\toprule[1.0pt]
Input Layer & A to D &W to A& A to P\\\hline
Middle& 0.437&0.594&0.474\\\hline
Last w/o Softmax&0.215&0.388&0.301\\\hline
\rowcolor{Gray}
\bf{Last w Softmax}&\bf{0.163}&\bf{0.348}&\bf{0.285}\\
\bottomrule[1.0pt]
\end{tabular}}
\end{center}
\vspace{-5mm}
\caption{\small {\bf Analysis of input features.} Within-class variance normalized by variance of all samples. Features of the last layer after softmax show the smallest variance within-class. }
\label{tb:feature_analysis}
\vspace{-3mm}

\end{table}

\begin{figure}
    \centering
    \includegraphics[width=\linewidth]{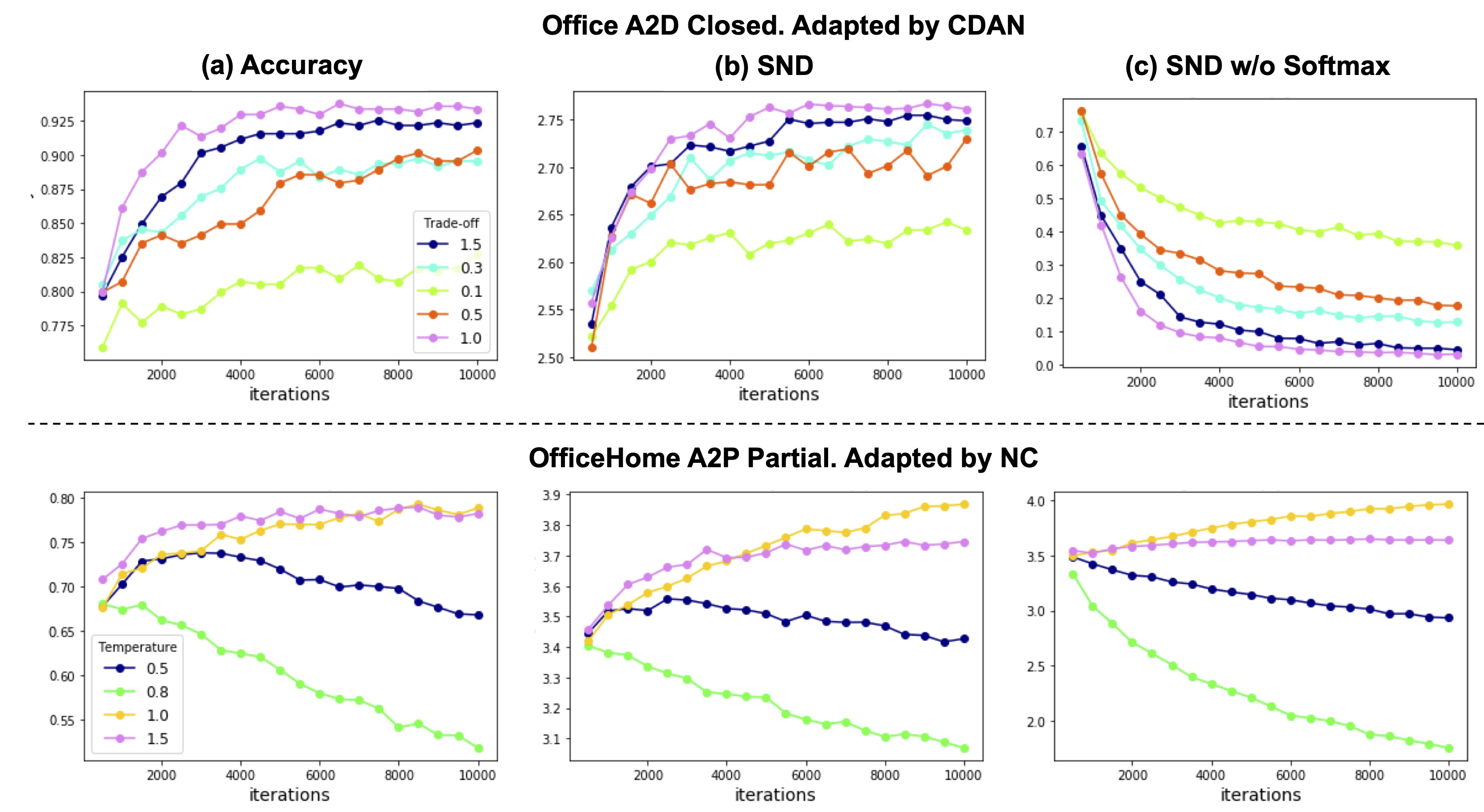}
    \vspace{-7mm}
    \caption{{\small {\bf Analysis of softmax features.} We remove the softmax layer to obtain target features and compute \textit{\oursv w/o Softmax}. The accuracy and \textit{\oursv w/o Softmax} have correlation, but the correlation depends on adaptation methods. Applying softmax makes the correlation consistent across methods. }}
    \label{fig:softmax_analysis}
    \vspace{-15pt}
\end{figure}

\textbf{Source Domain Selection.}\quad
We examine whether \oursv can select the best source domain given a target domain using the OfficeHome dataset. The task is to predict the best source domain from 3 candidates given a target domain. For simplicity, we do not use any adaptation method, thus we just train a model using source samples and evaluate using unlabeled target samples. As shown in Fig.~\ref{tb:source_select}, though \oursv does not always predict the best source domain, it always returns a model with upper-bound level performance.

\subsection{Analysis}
 \vspace{-1mm}

\textbf{Effectiveness of Softmax Normalization of Features.}\quad
Here, we analyze the effect of using softmax normalized features to compute $S_{ij}$ in Sec.~\ref{sec:wind}. We compute the relative within-class variance, \ie, within-class variance divided by the variance of all classes, and compare between different features in Table~\ref{tb:feature_analysis}.
The features we employ (\textit{Last w Softmax}) show the smallest relative variance, \ie, they separate each sample from other classes the best. Therefore, using the features allows us to ignore samples of other classes effectively in computing Eq. ~\ref{eq:distribution}. 

Next, we track \oursv without softmax normalization $S_{ij}$ in  Fig.~\ref{fig:softmax_analysis}.  
Note that we retain the softmax which normalizes the rows of the similarity matrix in Eq. ~\ref{eq:distribution}. 
The accuracy and \textit{\oursv w/o Softmax} have correlation, but the correlation depends on adaptation methods. The softmax normalization has the effect of highlighting difference between within-class and between class variance, which is a key to the success of \oursv. Other normalization methods, such as L2, didn't have the same effect.

\textbf{Possible Failure Cases and How to Avoid Them.}\quad
As we mention in the method section, first, if a model is not trained at all, the output does not characterize features of the target samples and \oursv does not work well. We can easily address this by monitoring the training loss on source samples. Second, one can also fool \oursv by training a model to collapse all target samples into a single point.
Empirically, we find that such a degenerate solution is hard to detect with any metrics including \oursv. One possible solution is to compare the feature visualizations of an adapted and an initial model.
We leave further analysis to future work. 

 \vspace{-2mm}
\section{Conclusion and Recommendations}
 \vspace{-2mm}
In this paper, we studied the problem of validating unsupervised domain adaptation methods, and introduced a novel criterion that considers how well target samples are clustered. Our experiments reveal a problem in existing methods' validation protocols. Therefore our recommendations to evaluate UDA algorithms in future research are as follows:
\begin{itemize}
 \vspace{-2mm}
\item Report which HPO method is used and describe the detail of validation if it includes hyper-parameters, \eg number of hold-out source samples.
 \vspace{-2mm}
\item A space to search hyper-parameters can be defined with the scale of loss and insight from previous works, but should be clearly discussed. 
 \vspace{-2mm}
\item Show the curve of the metric and accuracy. 
 \vspace{-2mm}
\item Publish implementations, including code for HPO.
 \vspace{-2mm}
\end{itemize}

It is important to design adaptation methods considering how HPO works on them. For example, methods requiring many hyper-parameters are hard to validate and, as we see in the right of Fig.~\ref{tb:source_select}, the difficulty of unsupervised validation differs from method to method.

Applying this protocol may reveal methods with highly performant upper-bounds on accuracy, but which are difficult to tune with any unsupervised validation criterion, including ours. In such scenarios, it may be reasonable to use a small set of target labels. However, this should be clearly discussed in the paper.


Finally, HPO is also crucial in open-set DA~\cite{busto2017open,saito2018open} and domain-adaptive object detection~\cite{dafaster}, and the topic of generalization in general. We leave extensions to these other tasks to future work. 

\textit{\bf{Acknowledgment.}}
\small{This work was supported by Honda, DARPA LwLL and NSF Award No. 1535797.}

\clearpage
\renewcommand{\thefigure}{\Alph{figure}}
 \renewcommand{\thetable}{\Alph{table}}
 \def\thesection{\Alph{section}}
\setcounter{section}{0}
\setcounter{figure}{0}
\setcounter{table}{0}

We first describe the details of the experiments. Then we show additional experimental results and analysis. 
\section{Experimental Details}
\textbf{Dataset.} In partial DA setting using OfficeHome, we choose the first 25 classes (in alphabetic order) as the target private classes following~\cite{cao2018partial}. In the experiments using DomainNet, we choose 126 classes out of 345 classes following~\cite{saito2019semi}. This is to remove classes that include some outlier objects or objects of multiple classes. 

\textbf{Implementation.} \quad As we mention in the main paper, we utilize official implementations to perform experiments. Specifically, we use the following implementations;
NC~\cite{saito2020universal}~\footnote[1]{\url{https://github.com/VisionLearningGroup/DANCE}}, 
CDAN~\cite{long2017conditional}~\footnote[2]{\url{https://github.com/thuml/CDAN}}, 
MCC~\cite{jin2020minimum}~\footnote[3]{\url{https://github.com/thuml/Versatile-Domain-Adaptation}}, ~
AdaptSeg~\cite{tsai2018learning}~\footnote[4]{\url{https://github.com/wasidennis/AdaptSegNet}}, and ~
ADVENT~\cite{vu2018advent}~\footnote[5]{\url{https://github.com/valeoai/ADVENT}}. 
We employ the configurations used by these implementations and tune the hyper-parameters described in the main paper. We will publish our implementation including these code. 

For NC~\cite{saito2020universal}, we tune the temperature parameter (Eq. 4 and 5 in ~\cite{saito2020universal}) used to compute similarity distribution. We multiple the $\tau$ with [0.5, 0.8, 1.0, 1.5] to find a optimal one.

\textbf{Semantic Segmentation.}  \quad
We describe the detail of an experiment on semantic segmentation.
First, we aim to tune a weight of trade-off between the source classification loss and domain confusion loss in this experiment. 

In AdaptSegNet~\cite{tsai2018learning}, the implementation defines two weights for two domain classifiers individually. We aim to tune a weight called ~\textit{lambda-adv-target1}. We aim to select the hyper-parameters from ($\lambda$ = 5.0 $\times$ $10^{-4}$, 3.0 $\times$ $10^{-4}$, 2.0 $\times$ $10^{-4}$, 1.0 $\times$ $10^{-4}$, 1.0 $\times$ $10^{-3}$, with 2.0 $\times$ $10^{-4}$ as its default setting). 
Similarly, for ADVENT~\cite{vu2018advent}, we aim to pick a trade-off parameter called \textit{LAMBDA-ADV-MAIN} from ($\lambda$ = 5.0 $\times$ $10^{-2}$ , 1.0 $\times$ $10^{-2}$, 1.0 $\times$ $10^{-3}$, 5.0 $\times$ $10^{-4}$, and 1.0 $\times$ $10^{-4}$, with 1.0 $\times$ $10^{-2}$ as its default setting). 

To compute source risk, we utilize 1,000 training source images as a source validation set and track mIoU over training iterations. 

\textbf{Toy Dataset.}  \quad
We utilize the implementation of DANN~\cite{ganin2016domain}~\footnote[6]{\url{https://github.com/GRAAL-Research/domain_adversarial_neural_network}}. We simply use their network architecture and other configurations. 
We generate source data from two Gaussian distributions with different means ((0,0) and (5,5)), which we regard as two classes. Then, we obtain target data by shifting the mean of one of the Gaussians. 

We will also publish the implementation modified for our experiment.  

\begin{figure*}
    \centering
    \includegraphics[width=\linewidth]{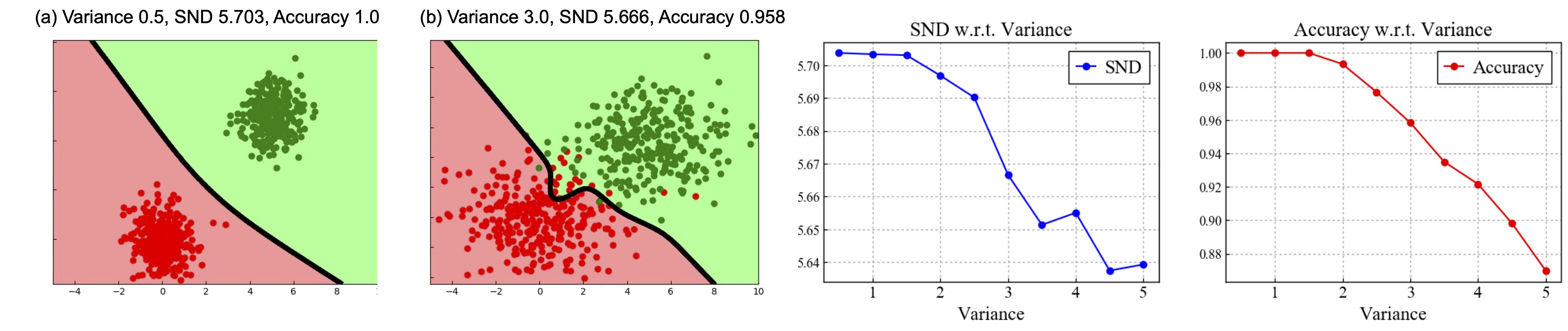}
    \vspace{-7mm}
    \caption{\oursv with respect to with-in class variance. Left (a)(b): Plots of changing the variance. We generate the data of two classes from two Gaussians with different means. As we show in the plot, we increase the variance of them while fixing their means. In this way, we observe the behavior of \oursv by the change of the feature density. Right: The change of \oursv and accuracy with respect to the variance. Since the concentration degree of features decreases with the increase of the variance, \oursv gets smaller with the increase.}
    \label{fig:toy_variance}
\end{figure*}

\section{Analysis in Toy Dataset}
Using the toy dataset, we show several characteristics of \oursv.
First, \oursv gets large if features have small within-class variance compared to the distance between classes. 
Second, \oursv outputs a large value when samples are generated from a single cluster. 

\textbf{\oursv and Within-class Variance.} \quad
\ours is designed to be large if the neighborhood samples are densely clustered, which means the feature variance within each cluster is small. We aim to confirm the relationship between \oursv and the within-class variance using the toy dataset. 
As shown in Fig.~\ref{fig:toy_variance}, we vary the variance of the Gaussian distributions that generate data of two classes while fixing their means. Note that we train and test a model on the same distribution since our goal here is to observe the behavior of \oursv for the different variance of features. The right of Fig.~\ref{fig:toy_variance} illustrates the result. As we expect, as the variance is increased, \oursv gets smaller. The accuracy also drops with the increase of the variance since the increase makes many hard-to-classify samples.

\textbf{\oursv and the Mode of the Data.} \quad
\begin{figure}
    \centering
    \includegraphics[width=\linewidth]{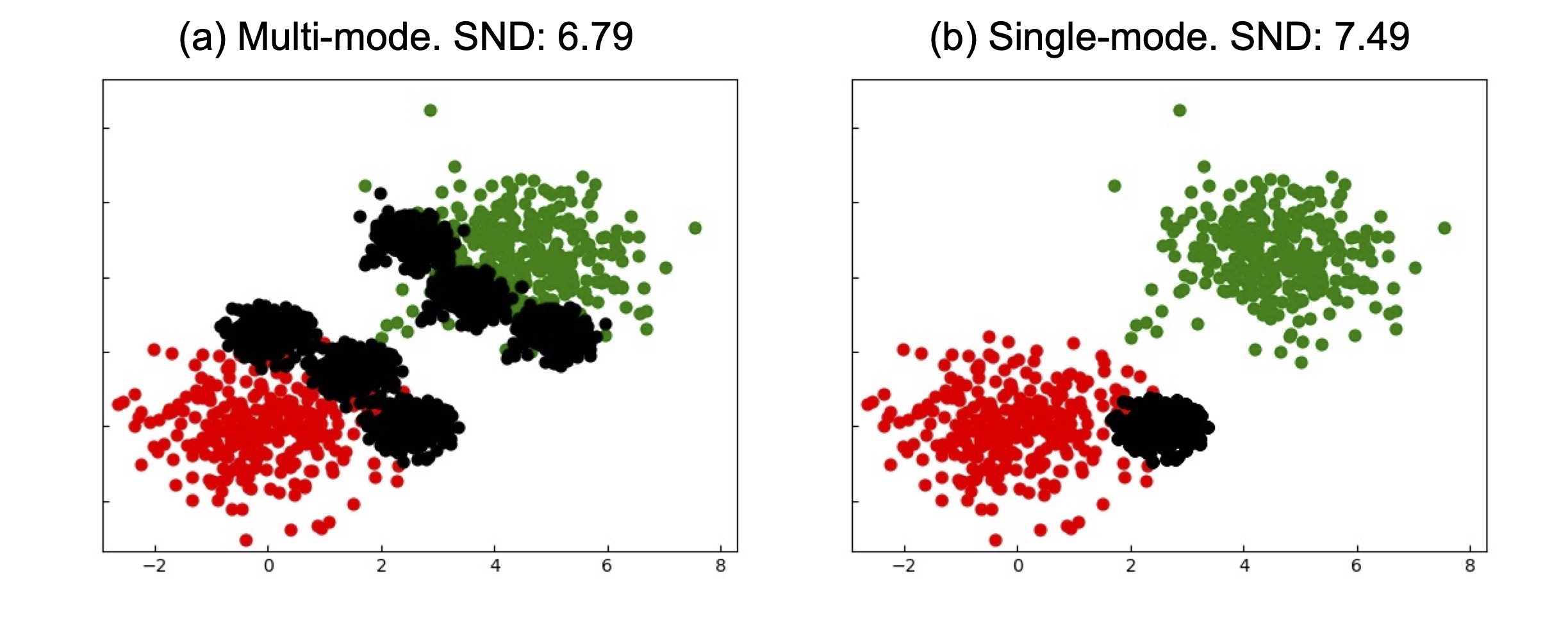}
    \vspace{-7mm}
    \caption{\oursv shows a large value for the data with a single mode. Black: Target samples. Red: Source samples of class 0. Green: Source samples of class 1. Left: Target samples are generated from 6 modes. Right: The same number of target samples as the left are generated from a single mode. \oursv of the right case is much larger than the left (7.49 vs 6.79).}
    \label{fig:toy_mode}
\end{figure}
In this experiment, we investigate the relationship between \oursv and the number of clusters in the target. Note that we assume we have a fixed number of target samples. 
As the number of clusters gets smaller, more target samples get similar since the total number of target samples is the same, and \oursv gets larger.
Fig.~\ref{fig:toy_mode} shows the result. In this experiment, we generate target data (Black dots in Fig.~\ref{fig:toy_mode}) by shifting the source distributions. In the left, we generate the target samples from 6 modes. In the right, the same number of target samples are generated from a single-mode. SND of the right case is much larger than that of the left (7.49 vs 6.79). 
If all target samples are from the green class, then SND picks the better model, but if they are actually from the red class, then SND picks the worse model. The failure case is hard to avoid as we discuss in Sec.~\ref{sec:analysis}.

\section{Additional Results}
We show results removed from our main paper due to limited space. 
\begin{figure*}
    \centering
    \includegraphics[width=\linewidth]{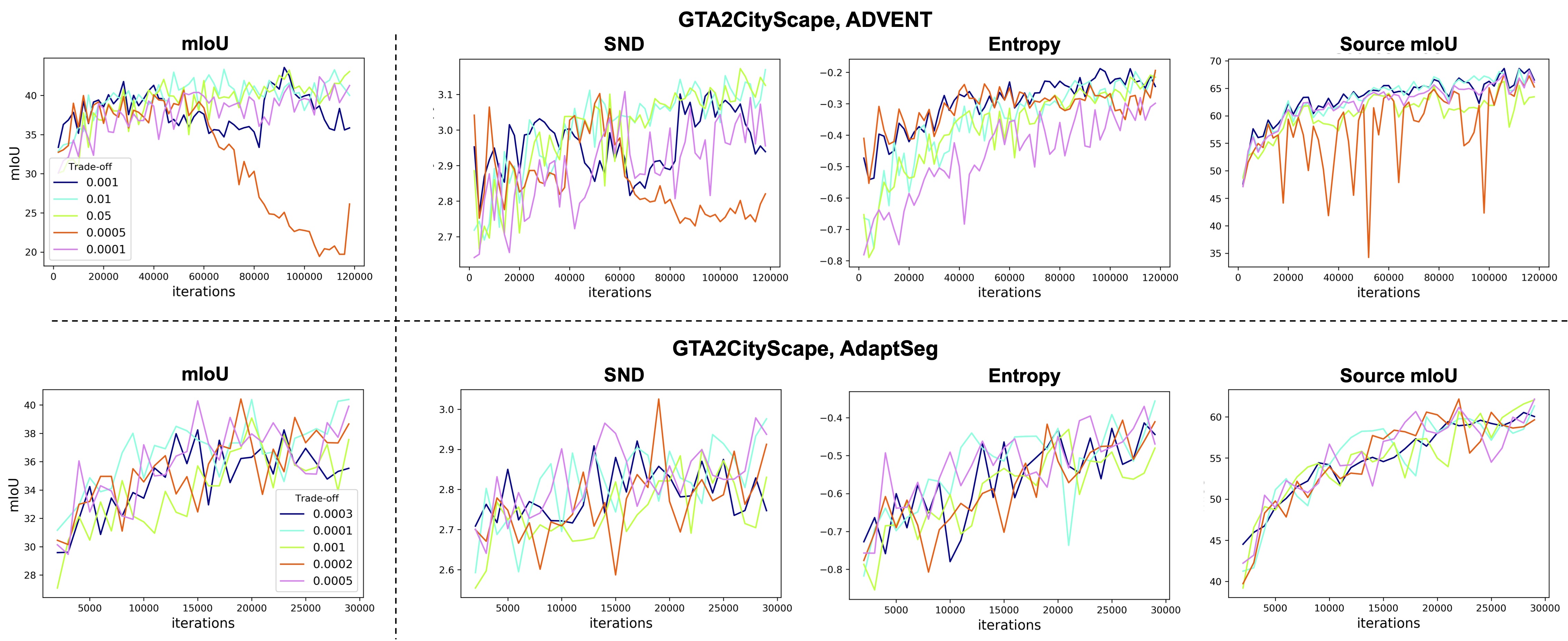}
    \caption{Semantic segmentation experiments (GTA5 to CityScaple) using AdaptSeg~\cite{tsai2018learning} and ADVENT~\cite{vu2018advent}. Different colors indicate different hyper-parameters. We validate the trade-off parameters between the source classification loss and domain-confusion loss. \oursv has a good correlation with mIoU (ground truth performance).}
    \label{fig:gta2city}
\end{figure*}

\textbf{Semantic Segmentation.} \quad
Fig.~\ref{fig:gta2city} shows iteration versus mIoU and HPO criterion. \oursv performs better than others on average in picking a better-adapted model (\ie, better target accuracy). Besides, we can observe that the performance of segmentation models is sensitive to hyper-parameters and training iterations.

\begin{figure*}
    \centering
    \includegraphics[width=\linewidth]{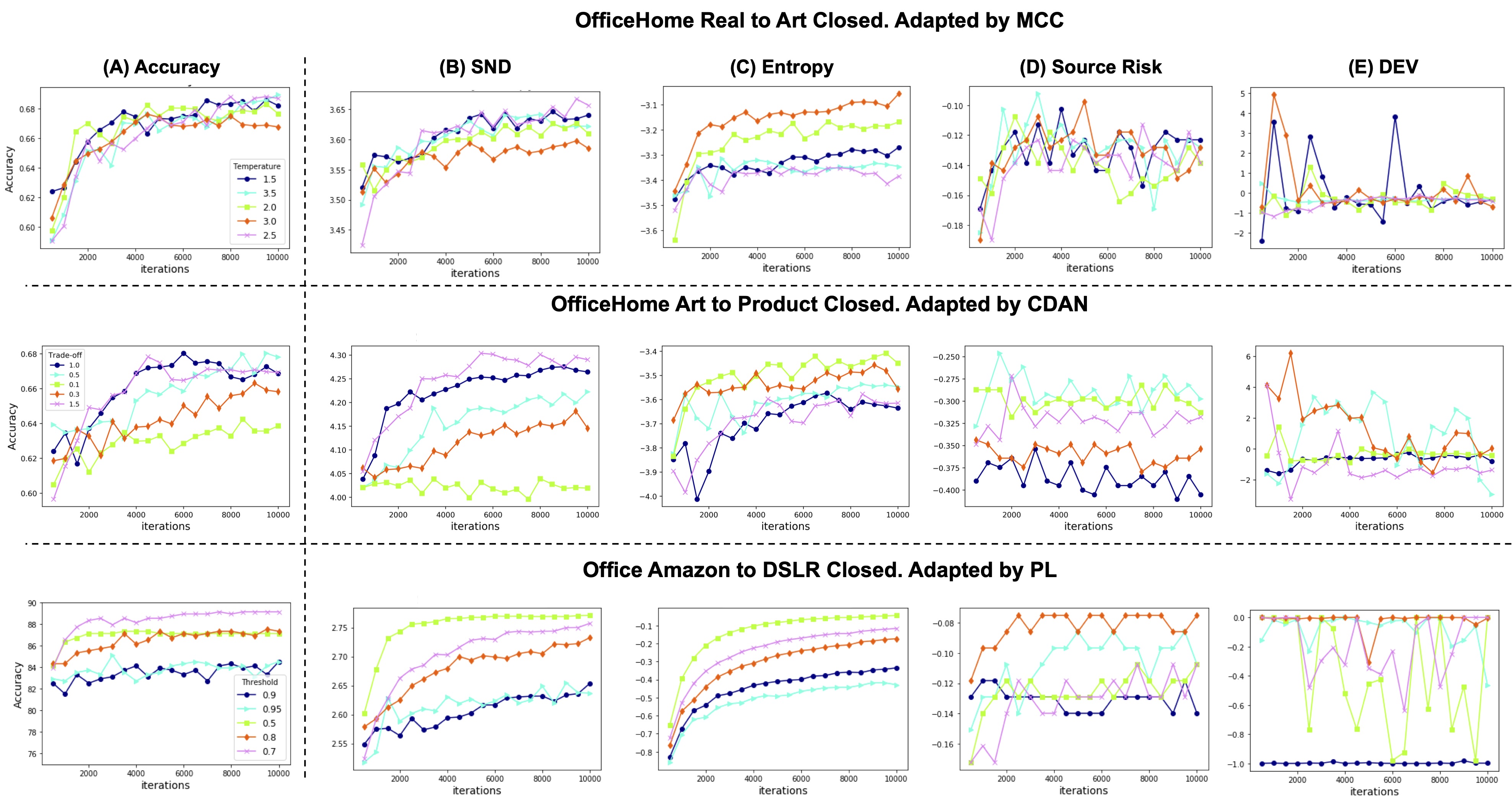}
    \vspace{-7mm}
    \caption{{\small Iteration versus accuracy and HPO criteria. Different colors indicate different hyper-parameters. To ease comparison between accuracy and criteria, we flip the sign of criteria for Entropy, Source risk, and DEV.}}
    \label{fig:graph_appendix}
    \vspace{-10pt}
\end{figure*}

\textbf{Image Classification.} \quad
Fig.~\ref{fig:graph_appendix} shows iteration versus accuracy and HPO criterion in image classification experiments. We show results of Pseudo-labeling (PL), CDAN~\cite{long2017conditional}, and MCC~\cite{jin2020minimum}.
\oursv performs better than others on average in picking a better-adapted model. Although \oursv does not always select the best model, \oursv shows a good correlation with accuracy. Entropy~\cite{morerio2017minimal} shows a similar behavior to \oursv in PL, but behaves in a totally different way in CDAN~\cite{long2017conditional}.

\begin{table*}[]
 \vspace{-3mm}
\begin{center}
\scalebox{1.0}{
    \begin{tabular}{l|cc|cc|ccc||c}
	\toprule[1.0pt]
	\multirow{2}{*}{Method}  &   \multicolumn{2}{c|}{Office} &\multicolumn{2}{c|}{OH CDA}  &\multicolumn{3}{c||}{OH PDA}&\multirow{2}{*}{Avg}      \\
 & A2D&W2A&R2A&A2P&R2A&A2P&P2C\\\hline

\bf{Lower Bound} &78.8 & 62.8 & 61.9 & 66.4 & 69.2 & 67.3 & 37.2 & 63.4 \\\hline
\bf{Source Risk}& \bf{81.3} & \bf{67.3} & 62.4 & 68.0 & 69.3 & 68.4 & 42.2 & 65.6 \\

\bf{DEV~\cite{you2019towards}}&\bf{81.3} & 66.2 & 65.3 & 67.8 & 70.4 & 70.2 & 44.2 & 66.5 \\

\rowcolor{Gray}
\bf{Entropy~\cite{morerio2017minimal}}&\bf{81.3} & \bf{67.3} & 62.8 &\bf{68.9} & 69.6 & 70.5 & 41.1 & 65.9 \\

\rowcolor{Gray}
\bf{\oursv(Ours)}&81.1 & \bf{67.3} & \bf{66.1} & 68.0 & \bf{72.2} & \bf{71.0} & \bf{44.2} & \bf{67.1} \\\hline

Upper Bound &84.7 & 68.9 & 66.9 & 68.9 & 72.2 & 71.3 & 45.3 & 68.3\\

	\bottomrule[1.0pt]
\end{tabular}}
\vspace{-3mm}
\caption{{\small \textbf{Results of MCD~\cite{saito2017maximum}.} \oursv performs the best on average. }}
\label{tb:mcd}\vspace{-20pt}

\end{center}

\end{table*}

\textbf{Results of MCD~\cite{saito2017maximum}.} \quad
We conduct experiments on tuning $\lambda$ of MCD~\cite{saito2017maximum}. MCD is a popular approach that employs the disagreement of two task-specific classifiers' output. As shown in the Table~\ref{tb:mcd}, \oursv shows the best performance on average. The result indicates the effectiveness of \oursv to tune classifier discrepancy-based adaptation methods.

\section{Additional Analysis}~\label{sec:analysis}
In this section, we show the detailed analysis of \ours and other criterion. 

\textbf{Failure Case.} \quad
\begin{figure*}
    \centering
    \includegraphics[width=\linewidth]{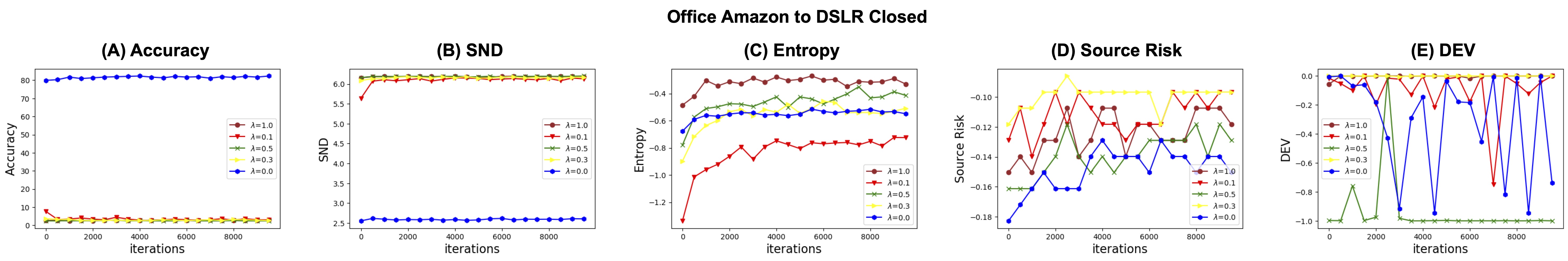}
    \caption{\textbf{Analysis of a possible failure case.} We train a network to correctly classify source samples and to classify all unlabeled target samples into one class. \textbf{Blue}: A model trained only with source classification loss. \textbf{Others}: Models trained to classify all target samples into a single class as well as trained to correctly classify source samples. Different colors indicate different weights, $\lambda$, for the target loss. No metric is able to identify the non-adapted model.}
    \label{fig:failure}
\end{figure*}
In Sec 4.4 in the main paper, we explain possible failure cases: One can fool \oursv by training a model to collapse all target samples into a single point. We analyze the behavior of metrics in this setting. Specifically, we train a network to correctly classify source samples and to classify all unlabeled target samples into one class. We call the models \textit{degenerated models}. Note that we will not employ this kind of a degenerated model in reality, but we train the models just to see the behavior of metrics.
We vary $\lambda$ for the target loss and compare the model with a non-adapted model. Fig.~\ref{fig:failure} shows the accuracy and the behavior of each metric. Since the model is trained to move all target samples to a single class, \oursv of degenerated models gets much larger than that of a non-adapted model(Blue). Other metrics are also not useful to identify the best model. Interestingly, training degenerated models for target does not decrease the accuracy of the source domain ((D) Source Risk). This is probably because the representational power of neural networks is rich enough to learn both the degenerated solution for the target and a good solution for the source domain. One possible solution to this problem is to compare the feature visualizations of the degenerated and a non-adapted model. We leave further analysis to future work.

\begin{figure*}[h!]
    \centering
    \includegraphics[width=\linewidth]{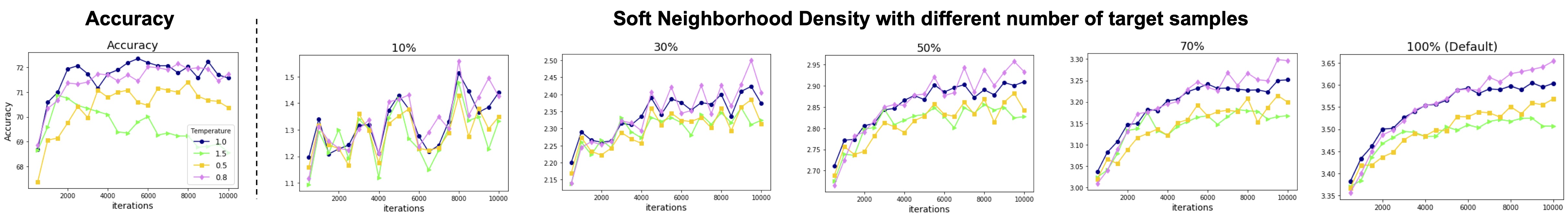}
    \vspace{-5mm}
    \caption{{\small Analysis of the number of target samples used to compute \oursv. Different colors indicate different hyper-parameters. We vary the number of target samples from $\frac{1}{10}N_t$ to $N_t$, where $N_t=2427$ is the number of target samples. We randomly sample the target samples. To reduce variance of \oursv, we need to sample certain number of target samples.}}
    \label{fig:target_numbers}
    \vspace{-10pt}
\end{figure*}

\textbf{Varying the Number of Target Samples.} \quad
We show analysis on the number of target samples necessary for \ours. 
Then, in the OfficeHome Real to Art closed adaptation, we employ NC~\cite{saito2020universal} and reduce the number of target samples used to calculate \oursv. We randomly sample a certain proportion of the target domain and compute \oursv. As shown in Fig.~\ref{fig:target_numbers}, \oursv is not very sensitive to the number of target samples. However, when we sample a small number of samples (10\% case), \ours becomes a little unstable. 
\begin{figure*}
    \centering
    \includegraphics[width=\linewidth]{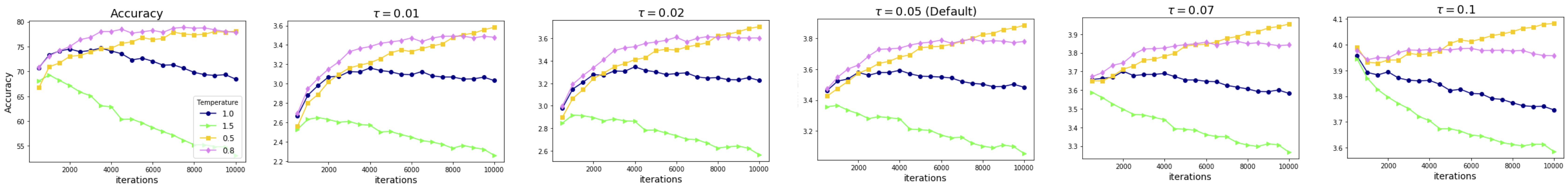}
    \vspace{-5mm}
    \caption{{\small Analysis of the temperature value used to compute \oursv. Different colors indicate different hyper-parameters. We vary the value of the temperature of Eq. 2, \ie, 0.01, 0.03, 0.05 (default), 0.07, 0.1. The result indicates that \oursv shows consistent results across different temperature values.}}
    \label{fig:temperature}
    \vspace{-10pt}
\end{figure*}

\textbf{Temperature Parameter.} \quad
We fix the temperature parameter ($\tau$) in Eq. 2 (See our main draft.) as 0.05 in all of our experiments. 
Then, in the OfficeHome Art to Product partial domain adaptation, we employ NC~\cite{saito2020universal} and vary the value of $\tau$. In Fig.~\ref{fig:temperature}, we compare the resulting curve of \oursv with the accuracy curve. We have two observations: \oursv is not very sensitive to the value of $\tau$ in selecting the best model; but, the large temperature can make \oursv inconsistent with the accuracy as the rightmost ($\tau=0.1$) result indicates. This result indicates the necessity of the temperature scaling. The scaling enables to ignore samples embedded far away and to compute the density of neighbors. 
\begin{figure*}[h!]
    \centering
    \includegraphics[width=\linewidth]{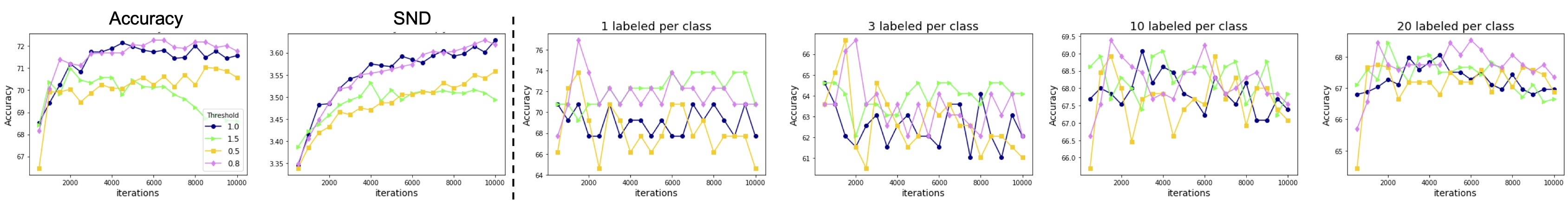}
    \vspace{-3mm}
    \caption{{\small Iteration versus accuracy and accuracy of the subset of a target domain. Different colors indicate different hyper-parameters. We subsample labeled target samples (1, 3, 10, 20 samples per class) and compute the accuracy. Many number of labeled samples is necessary to resemble the performance of a whole target domain.}}
    \label{fig:labeled_target}
    \vspace{-10pt}
\end{figure*}
\begin{figure*}[h!]
    \centering
    \includegraphics[width=0.9\linewidth]{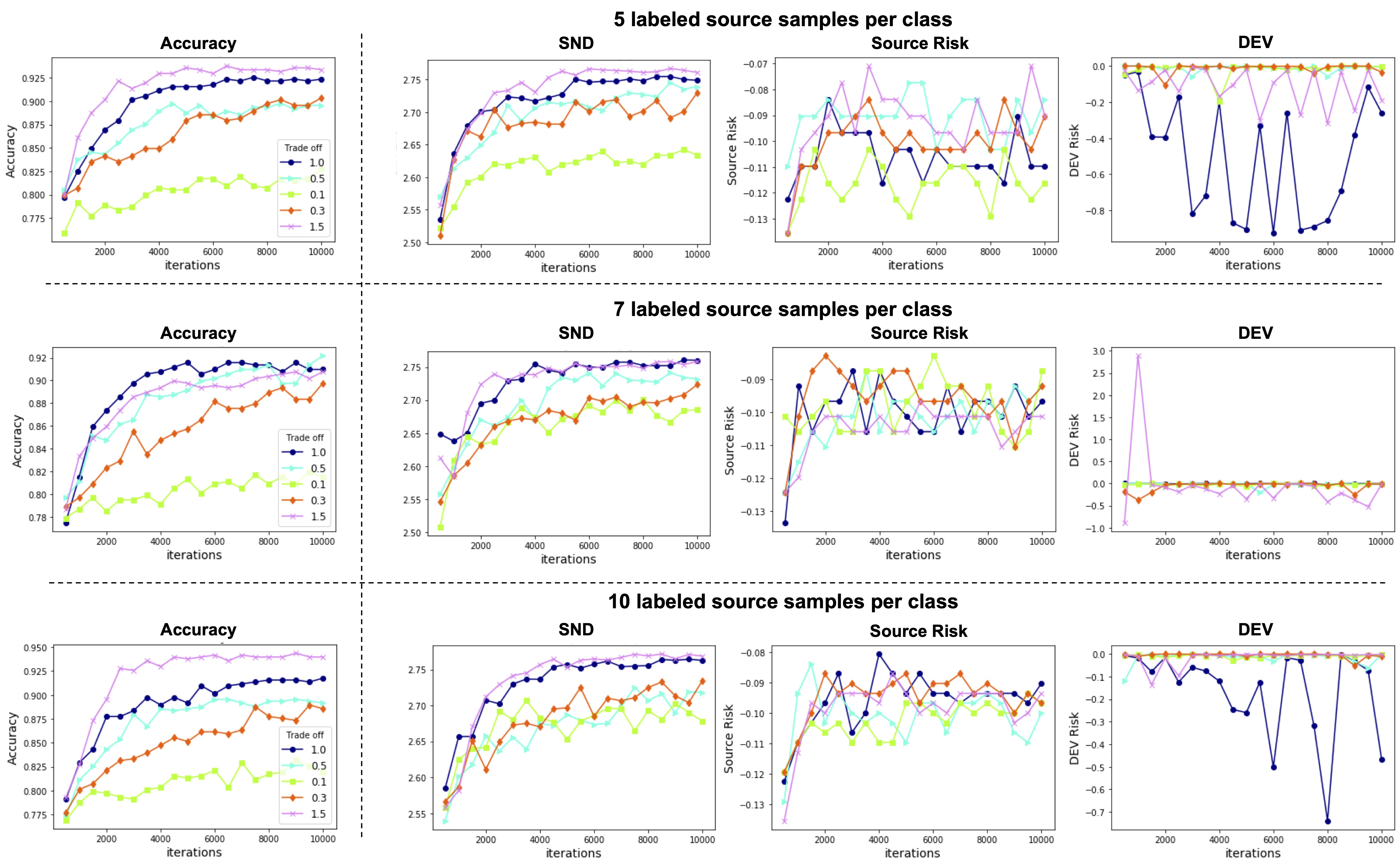}
    \vspace{-3mm}
    \caption{{\small Analysis of the number of labeled source samples used for validation. We vary the number of the labeled source samples to compute source risk and DEV risk. Different colors indicate different hyper-parameters. The result indicates that even though we increase the number of source validation samples, the risks are not reliable to select hyper-parameters.}}
    \label{fig:labeled_source}
    \vspace{-10pt}
\end{figure*}

\textbf{\ours Versus Validation with a Few Labeled Target Samples.} \quad
Some papers propose to utilize a few labeled target samples to tune hyper-parameters. Although the way of tuning violates the assumption of UDA, we investigate how well the criterion is effective to pick a good hyper-parameter in Fig.~\ref{fig:labeled_target}. We employ the OfficeHome Real to Art closed adaptation using NC~\cite{saito2020universal}. 
We increase the number of validation target samples per class from 1 to 20 and compare the result with \oursv. When the number of labeled target samples is small, the validation accuracies are not stable and have high variance. To obtain stable and reliable results, we need to have many labeled target samples whereas \oursv is an unsupervised criterion and shows reliable results. In a real application, having a few labeled target samples may not be always hard as stated in \cite{saito2019semi}. However, as this result indicates, monitoring only the accuracy of few samples may not provide a good model. Even in such a setting, combining \oursv will be a good way to tune hyper-parameters.

\textbf{Analysis of the Number of Source Validation Samples on Source Risk and DEV~\cite{you2019towards}.} \quad
We further analyze the cause of failures of source risk and DEV~\cite{you2019towards}. We increase the number of labeled source samples and observe the behavior of two criteria. We use the Amazon to DSLR setting adapted by CDAN~\cite{long2017conditional}. Even when we use a large proportion of source samples as a validation set (We utilize more than 10 \% of source samples in the case of 10 labeled samples per class.), the two criteria are not well correlated with the accuracy of the target domain. This result indicates the using source risk is limited to choosing good hyper-parameters.

{\small
\bibliographystyle{ieee_fullname}
\bibliography{egbib}
}

\end{document}